\documentclass[conference]{IEEEtran}
\usepackage{times}

\usepackage[numbers]{natbib}
\usepackage{multicol}
\usepackage[bookmarks=true]{hyperref}
\usepackage{amsfonts, amsmath, amssymb}
\usepackage{todonotes}
\usepackage{booktabs}
\usepackage{float}

\usepackage{etoolbox}
\makeatletter
\patchcmd{\@makecaption}
  {\scshape}
  {}
  {}
  {}
\makeatletter
\patchcmd{\@makecaption}
  {\\}
  {.\ \raggedright }
  {}
  {}
\makeatother
\def\tablename{Table}

\begin{document}

\title{Augmenting Differentiable Simulators with\\ Neural Networks to Close the Sim2Real Gap}




%
\author{
\authorblockN{
Eric Heiden\authorrefmark{1},
David Millard\authorrefmark{1},
Erwin Coumans\authorrefmark{2}, and
Gaurav S. Sukhatme\authorrefmark{1}
}

\authorblockA{
\authorrefmark{1}%
Robotics Embedded Systems Lab\\
University of Southern California,
Los Angeles, California 90089\\
Email: \texttt{\{%
\href{mailto:heiden@usc.edu}{heiden},%
\href{mailto:dmillard@usc.edu}{dmillard},%
\href{mailto:gaurav@usc.edu}{gaurav}%
\}@usc.edu}
}

\authorblockA{
\authorrefmark{2}%
Google, Mountain View, California 94043\\
Email: \texttt{\href{mailto:erwincoumans@google.com}{erwincoumans@google.com}}
}
}

\maketitle

\begin{abstract}
We present a differentiable simulation architecture for articulated rigid-body dynamics that enables the augmentation of analytical models with neural networks at any point of the computation. Through gradient-based optimization, identification of the simulation parameters and network weights is performed efficiently in preliminary experiments on a real-world dataset and in sim2sim transfer applications, while poor local optima are overcome through a random search approach.
\end{abstract}

\IEEEpeerreviewmaketitle

\section{Introduction and Related Work}

\begin{figure*}[ht!]
    \centering
    \includegraphics[height=4.5cm]{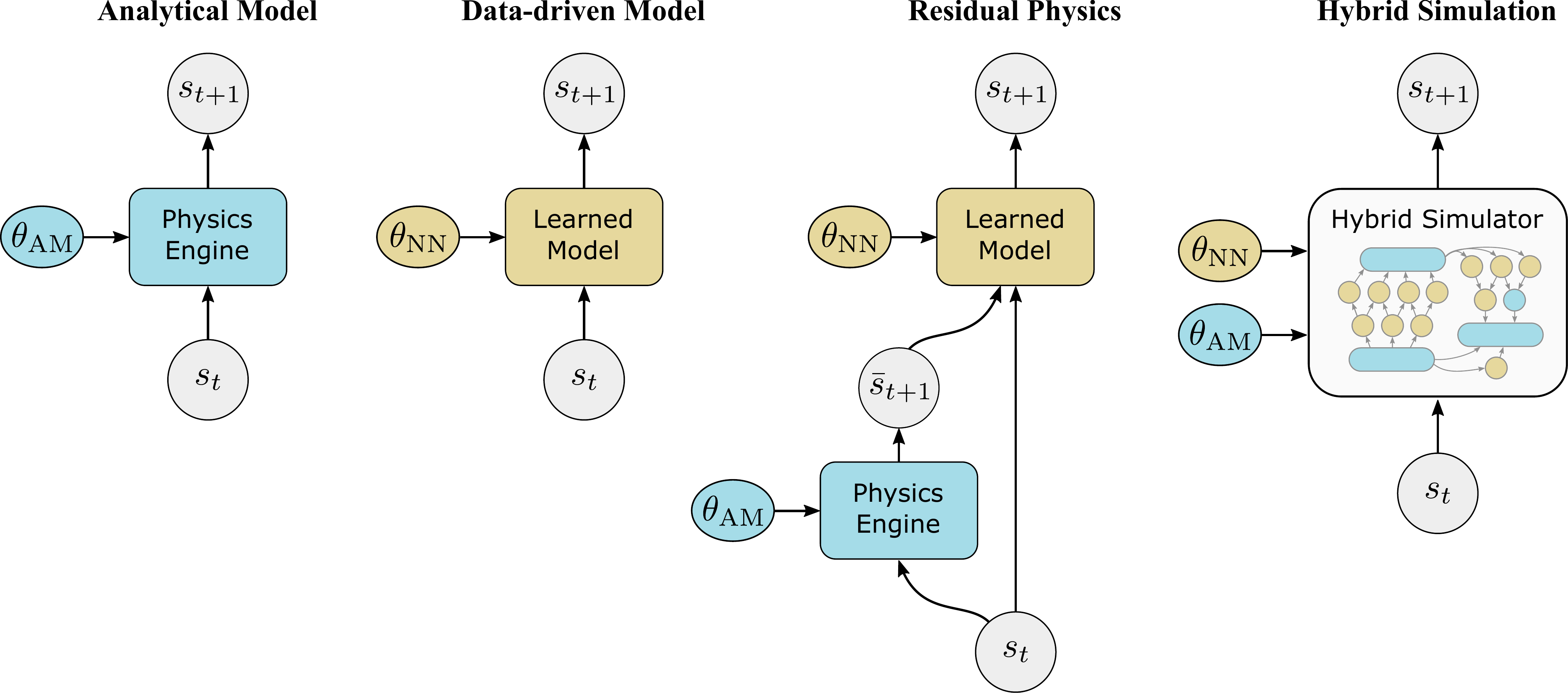}\hfill
    \includegraphics[height=4.5cm]{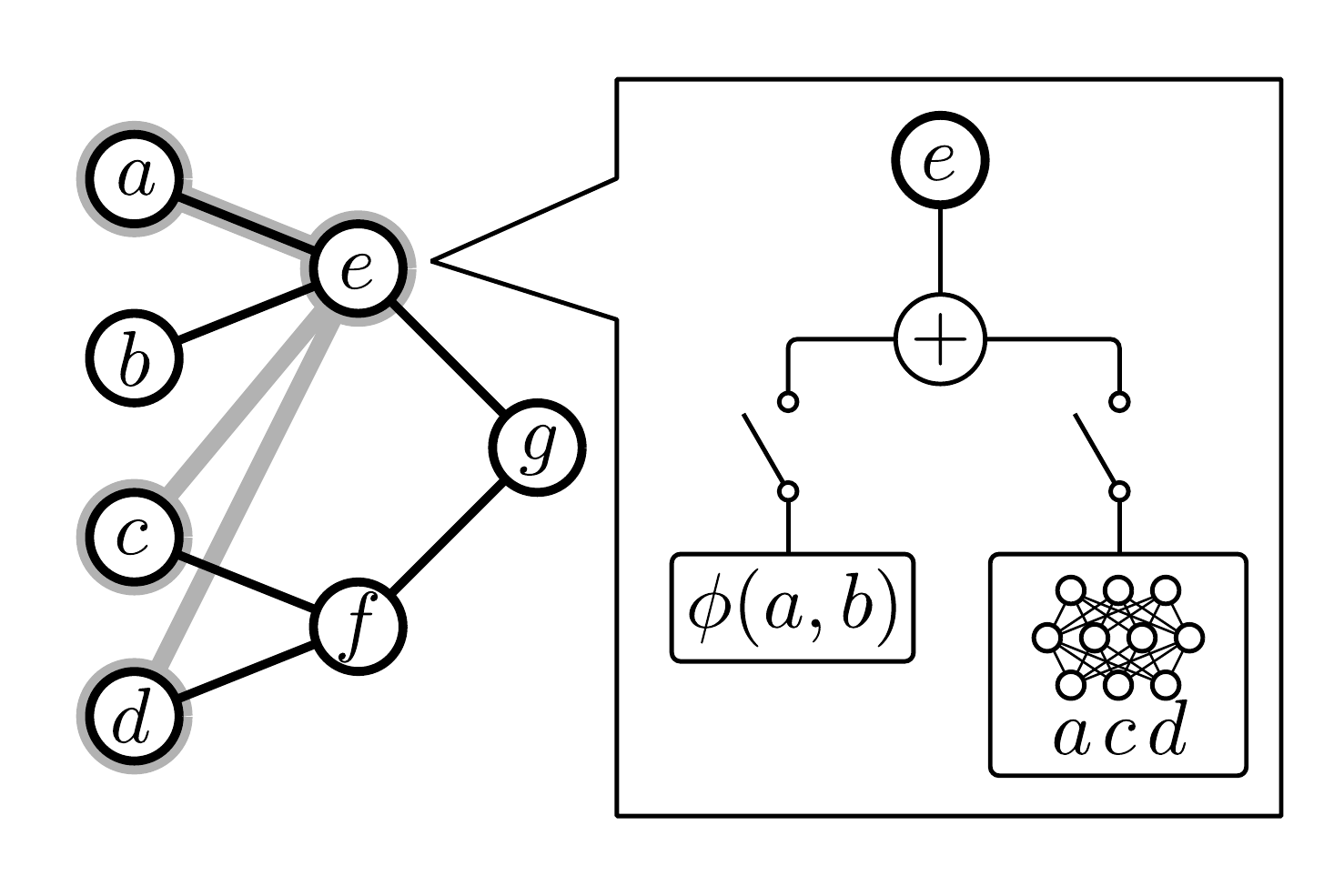}
    \caption{\textit{Left:} comparison of various model architectures (cf.~Anurag et al.~\cite{anurag2018hybrid}). \textit{Right:} augmentation of differentiable simulators with our proposed neural scalar type where variable $e$ becomes a combination of an analytical model $\phi(\cdot,\cdot)$ with inputs $a$ and $b$, and a neural network whose inputs are $a$, $c$, and $d$.}
    \label{fig:approaches}
\end{figure*}

Simulators are crucial tools for planning and control algorithms to tackle difficult real world robotics problems. In many cases, however, such models diverge from reality in important ways, leading to algorithms that work well in simulation and fail in reality. Closing the sim2real gap has gained significant interest, and various dynamics modeling approaches have been proposed (\autoref{fig:approaches} left).

Various methods learn system dynamics from time series data of a real system. Such ``intuitive physics'' models often use deep graph neural networks to discover constraints between particles or bodies~\cite{battaglia2016interaction, xu2019physics, he2019physics, raissi2018physics, mrowca2018physics, chen2018neural, li2018learning, sanchezgonzalez2020learning}.
We propose a general-purpose hybrid simulation approach that combines analytical models of dynamical systems with data-driven residual models that learn parts of the dynamics unaccounted for by the analytical simulation models.

Originating from traditional physics engines~\cite{coumans2013bullet,todorov2012mujoco,lee2018dart}, differentiable simulators have been introduced that leverage automatic, symbolic or implicit differentiation to calculate parameter gradients through the analytical physics models for rigid-body dynamics~\cite{giftthaler2017autodiff,carpentier2018analytical,peres2018lcp, koolen2019rbd-julia, heiden2019ids, heiden2019real2sim}, light propagation~\cite{NimierDavidVicini2019Mitsuba2, heiden2020lidar}, and other phenomena~\cite{hu2020difftaichi,liang2019differentiable}.

Residual physics models~\cite{zeng2019tossingbot,anurag2018hybrid,hwangbo2019learning,golemo2018neuralaugsim} augment physics engines with learned models to reduce the sim2real gap. Most of them introduce residual learning applied to the output of the physics engine, while we propose a more fine-grained approach, similar to Hwangbo et al.~\cite{hwangbo2019learning}, where only at some parts in the simulator data-driven models are introduced. While in~\cite{hwangbo2019learning} the network for actuator dynamics is trained through supervised learning, our end-to-end differentiable model allows backpropagation of gradients from high-level states to any part of the computation graph, including neural network weights, so that these parameters can be optimized efficiently, for example from end-effector trajectories.

\begin{figure}
    \centering
    \includegraphics[height=3.5cm, clip, trim=3cm 1cm 1cm 2cm]{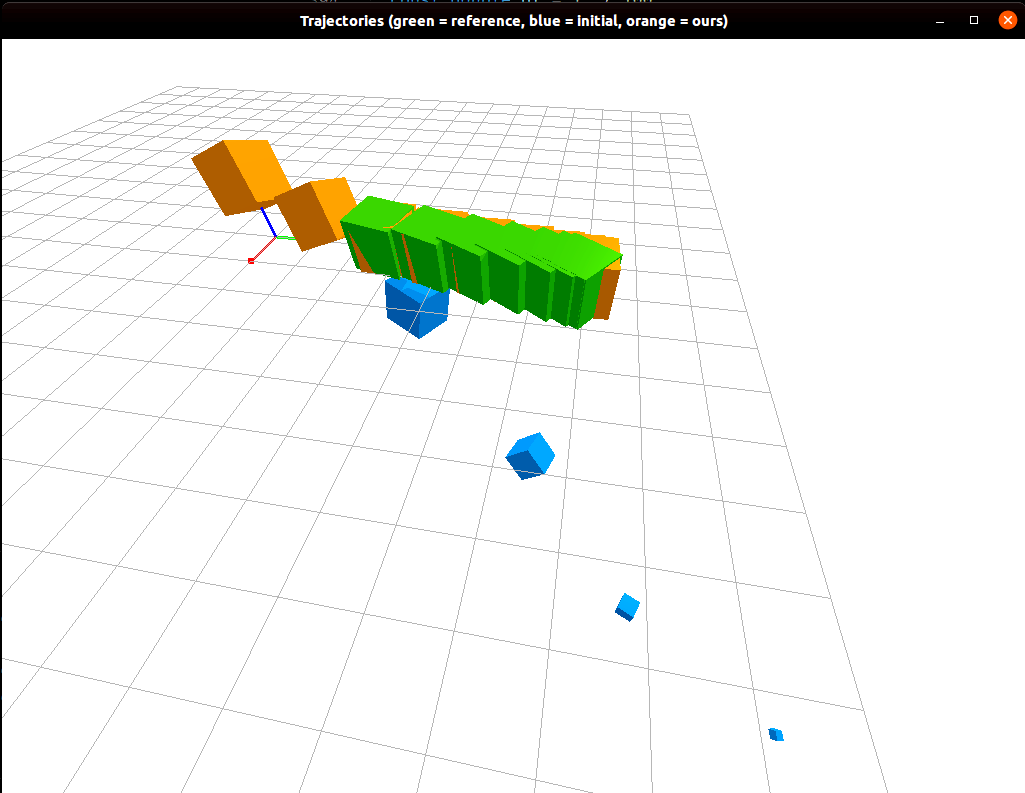}\hfill
    \includegraphics[height=4.5cm]{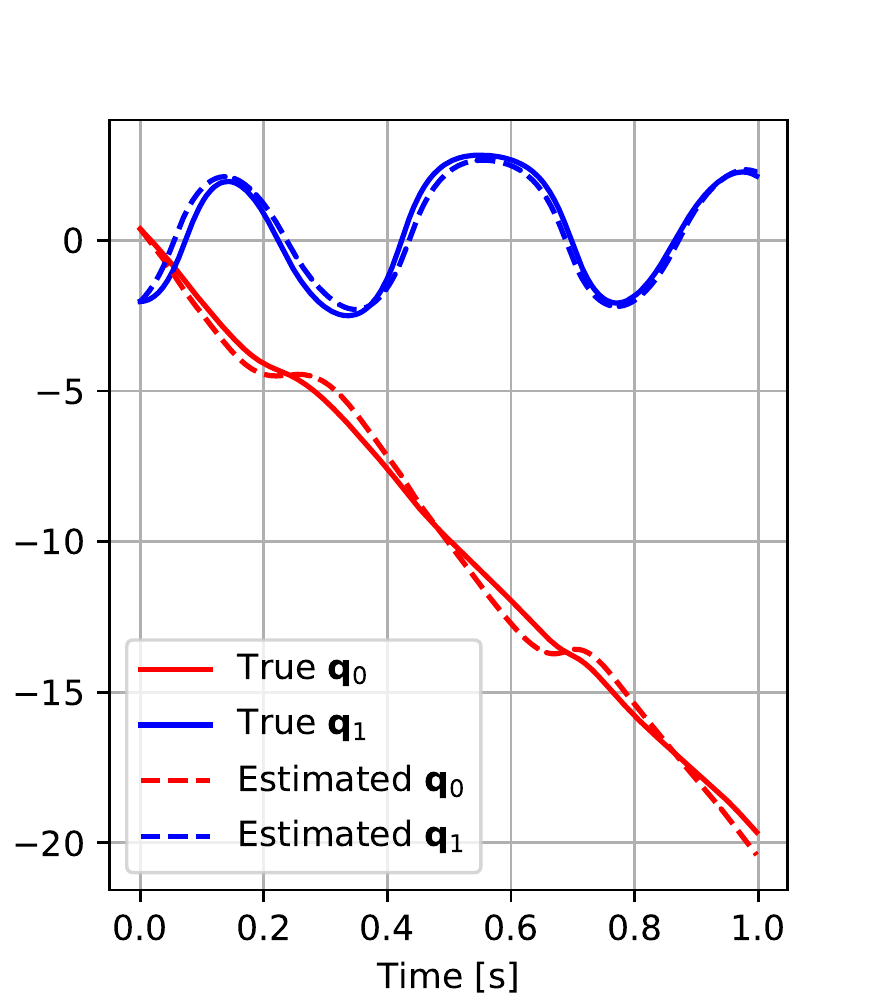}
    \caption{\emph{Left:} Trajectories from rigid body contact simulation of a cube thrown to the right. Starting with poor model parameters, the box falls through the ground (blue). After optimizing \autoref{eq:sim2rel_loss}, our simulation (orange) closely matches the target trajectory (green). \emph{Right:} After system identification of a real double pendulum~\cite{asseman2018learning}, the sim2real gap is strongly reduced.}
    \label{fig:results}
\end{figure}

\vspace*{-1cm}

\section{Approach}
We propose a technique for hybrid simulation that leverages differentiable physics models and neural networks to allow for efficient system identification, design optimization, and gradient-based trajectory planning. By enabling any part of the simulation to be replaced or augmented by neural networks, we can learn unmodeled effects from data. Through template meta-programming, our open-source C++ implementation\footnote{\url{https://github.com/google-research/tiny-differentiable-simulator}} allows any variable participating in the simulation to be augmented by neural networks that accept input connections from any other variable. In the simulation code, such \emph{neural scalars} (\autoref{fig:approaches} right) are assigned a unique name, so that in a separate experiment code a ``neural blueprint'' is defined that declares the neural network architectures and sets the network weights. We compute gradients of the weights and analytical simulation parameters using the multi-dimensional dual number implementation from Ceres~\cite{ceres-solver} and have support for many other automatic differentiation libraries.

\begin{table}[]
    \resizebox{\columnwidth}{!}{%
    \begin{tabular}{p{3.8cm}cccc}
    \toprule
    & \bf Analytical & \bf Data-driven & \bf End2end $\nabla$ & \bf Hybrid \\\midrule
    Physics engine~\cite{coumans2013bullet,todorov2012mujoco,lee2018dart} & \checkmark \\
    Residual physics~\cite{hwangbo2019learning,anurag2018hybrid,zeng2019tossingbot,golemo2018neuralaugsim} & \checkmark & \checkmark & & \checkmark \\
    Learned physics~\cite{sanchez2020learning,battaglia2016interaction,li2018learning,jiang2018datacontact} & & \checkmark & \checkmark \\
    Differentiable sim.~\cite{giftthaler2017autodiff,hu2020difftaichi,carpentier2018analytical,peres2018lcp} & \checkmark & & \checkmark \\
    Ours & \checkmark & \checkmark & \checkmark & \checkmark \\
\bottomrule
    \end{tabular}
    }
    \caption{Comparison of dynamics modeling approaches (only selected works) along the axes of analytical and data-driven modeling, end-to-end differentiability, and hybrid approaches.}
    \label{tab:comparison}
\end{table}

\section{System Identification}

Given the state trajectory $\{s^*_t\}_{t=1}^T$ from the target system, we optimize the following loss for system identification:
\begin{align}
\label{eq:sim2rel_loss}
\underset{\theta = [\theta_{AM}, \theta_{NN}]}{\operatorname{minimize}} ~~ \mathcal{L} = \sum_t ||f_\theta(s_{t-1}) - s^*_t||^2 + R||\theta_{NN}||^2,
\end{align}
where $f_\theta(\cdot)$ is the discrete dynamics function mapping from the previous simulation state $s_{t-1}$ to the current state $s_t$ which is implemented by our physics engine given the parameter vector $\theta$ that consists of the parameters $\theta_{AM}$ for the analytical model, plus the parameters $\theta_{NN}$ that correspond to the weights of the neural networks in the simulation. To ensure the residual dynamics learned by the neural networks are minimal, we regularize the network weights by factor $R$ which penalizes large state contributions.

\section{Overcoming Local Optima}
We solve the nonlinear least squares problem from \autoref{eq:sim2rel_loss} using the Levenberg-Marquardt algorithm (LMA). Such a gradient-based optimization method quickly finds local optima, but due to the highly nonconvex loss landscapes commonly encountered in system identification problems for nonlinear dynamics, the resulting parameter estimates often exhibit a poor fit to the real world data. To escape such poor local minima, we adapt a random search strategy, \emph{parallel basin hopping} (PBH)~\cite{mccarty2018parallel}, that, in our instantiation, runs multiple LMA solver and simulation instances in parallel while continuously randomizing the initial parameters from which the local solvers are restarted after convergence criteria, time limits, or maximum iteration counts are met. 

\section{Results}
We present preliminary results for sim2sim transfer to match the hybrid dynamics model to richer analytical simulations. Additionally, we demonstrate our system identification approach on a real-world dataset.

In our first experiment, we transfer rigid-body contact dynamics simulated using a velocity-level contact model formulated as a linear complementarity problem~\cite{anitescu1997formulating}. Our hybrid simulator uses a point-based nonlinear-spring contact model where the normal force is solved analytically through the Hunt-Crossley model~\cite{hunt1975coefficient} and the friction force is learned by a neural network that receives the relative velocities, contact normal force and penetration depth as input. Before optimizing the analytical and neural model parameters, the trajectories of a cube thrown horizontally on the ground differ dramatically. After system identification using PBH applied on \autoref{eq:sim2rel_loss} given trajectories of positions and velocities from the target system, the gap is significantly reduced (\autoref{fig:results} left).

In the next experiment, we apply our approach to a real-world dynamical system. Given joint position trajectories from the double-pendulum dataset provided by Asseman et al.~\cite{asseman2018learning}, we optimize inertia, masses, and link lengths of our simulated model and achieve a minimal sim2real gap (\autoref{fig:results} right).

\section{Conclusion} 
\label{sec:conclusion}

We have demonstrated a simulation architecture that allows us to insert neural networks at any place in a differentiable physics engine to augment analytical models with the ability to learn dynamical effects from data. In our preliminary experiments, efficient gradient-based optimizers quickly converge to simulations that closely follow the observed trajectories from the target systems, while poor local minima are overcome through a random search strategy.

Future research is directed towards more automated ways to identify where such extra degrees of freedom are needed to close the sim2real gap given a few trajectories from the real system. Our loss function in \autoref{eq:sim2rel_loss} regularizes the contributions of the neural networks to the overall system dynamics. Nonetheless, this approach does not prevent violating basic laws of physics, such as energy and momentum preservation. Hamiltonian~\cite{greydanus2019hamiltonian} and (Deep) Lagrangian neural networks~\cite{lutter2019delan,cranmer2020lagrangian} explicitly constrain the solution space to remain consistent with such principles but need to be further investigated in the context of residual models in hybrid simulators.



\bibliographystyle{plainnat}
\bibliography{references}

\end{document}